\documentclass{article} 
\usepackage{nips13submit_e,times}
\usepackage{hyperref}
\usepackage{url}
\usepackage{amsmath, amsthm, amssymb}
\usepackage{graphicx}
\usepackage{caption}
\usepackage{subcaption}

\newcommand{\refSection}[1]{Section~\ref{#1}}
 
\newcommand{\refCite}[1]{\cite{#1}}

\newcommand{\refEq}[1]{eq(\ref{#1})}

\newcommand{\ie}[0]{\textit{i.e.},~}
\newcommand{\eg}[0]{\textit{e.g.},~}
\newcommand{\aka}[0]{\textit{a.k.a}.~}

\newcommand{\xv}[0]{\ensuremath{\mathbf{x}}}
\newcommand{\yv}[0]{\ensuremath{\mathbf{y}}}

\newcommand{\vv}[0]{\ensuremath{\mathbf{v}}}
\newcommand{\vvd}[1]{\ensuremath{\mathbf{v}^{(#1)}}}
\newcommand{\vd}[1]{\ensuremath{{v}^{(#1)}}}
\newcommand{\hv}[0]{\ensuremath{\mathbf{h}}}
\newcommand{\av}[0]{\ensuremath{\mathbf{a}}}
\newcommand{\bv}[0]{\ensuremath{\mathbf{b}}}
\newcommand{\W}[0]{\ensuremath{\mathbf{W}}}
\newcommand{\thetaeps}[0]{\ensuremath{\widetilde{\theta}}}
\newcommand{\aveps}[0]{\ensuremath{\widetilde{\mathbf{a}}}}
\newcommand{\bveps}[0]{\ensuremath{\widetilde{\mathbf{b}}}}
\newcommand{\Weps}[0]{\ensuremath{\widetilde{\mathbf{W}}}}
\newcommand{\aeps}[0]{\ensuremath{\widetilde{a}}}
\newcommand{\beps}[0]{\ensuremath{\widetilde{b}}}

\newcommand{\Eeps}[0]{\ensuremath{\widetilde{E}}}
\newcommand{\xvh}[0]{\ensuremath{\widehat{\mathbf{x}}}}

\newcommand{\set}[1]{\ensuremath{\mathcal{#1}}}

\newcommand{\p}{\ensuremath{\mathbb{P}}}
\newcommand{\peps}{\ensuremath{\widetilde{\mathbb{P}}}}

\newcommand{\ex}[2]{\ensuremath{{\mathbb{E}}_{#1}{\left [ \; #2  \; \right ]}}}
\newcommand{\Z}{\ensuremath{{Z}}}

\newtheorem{theorem}{Theorem}

\newtheorem{lemma}[theorem]{Lemma}

\title{Training Restricted Boltzmann Machine by Perturbation}

\author{
Siamak Ravanbakhsh, Russell Greiner \\
Department of Computing Science\\
University of Alberta\\
\texttt{\{mravanba,rgreiner@ualberta.ca\} } \\
\And
Brendan J. Frey \\
Prob. and Stat. Inf. Group \\
Universi of Toronto\\
\texttt{frey@psi.toronto.edu} \\
}

\nipsfinalcopy 

\begin{document}

\maketitle

\begin{abstract}
A new approach to maximum likelihood learning of discrete graphical models and RBM in particular is introduced.
Our method, Perturb and Descend (PD) is inspired by two ideas (I) perturb and MAP method for sampling
 (II) learning by Contrastive Divergence minimization. In contrast to perturb and MAP, 
 PD leverages training data to learn the models that do not allow efficient MAP estimation. 
During the learning, to produce a sample from the current model, we start from a training data 
and descend in the energy landscape of the ``perturbed model'', 
for a fixed number of steps, or until a local optima is reached.
For RBM, this involves linear calculations and thresholding which can be very fast.
Furthermore we show that the amount of perturbation is closely related to the temperature parameter and it can regularize the model by producing robust features resulting in sparse hidden layer activation.
\end{abstract}

\section{Introduction}
The common procedure in learning a Probabilistic Graphical Model (PGM) is to maximize the likelihood of observed data,
by updating the model parameters along the gradient of the likelihood.
This gradient step requires inference on the current model, which may be performed using deterministic or  
a Markov Chain Monte Carlo (MCMC) procedure \refCite{koller_probabilistic_2009}. Intuitively, the gradient step attempts to 
update the parameters to \emph{increase} the unnormalized probability of the observation, while
\emph{decreasing} the sum of unnormalized probabilities over all states --\ie the partition function.
The first part of the update is known as positive phase and the second part is referred to as the negative phase.
An efficient alternative is Contrastive Divergence (CD) \refCite{hinton2002training} training, in which 
the negative phase only decreases the probability of the configurations that are in the vicinity of training data. 
In practice these neighboring states are sampled by taking few steps on Markov chains that are initialized by training data.

Recently Perturbation methods combined with efficient Maximum A Posteriori (MAP) solvers were used to efficiently sample PGMs \refCite{papandreou2010gaussian,papandreou2011perturb,hazan2013sampling}.
Here a basic idea from extreme value theory is used, which states that the MAP assignments for particular perturbations of any Gibbs distribution can replace unbiased samples from the unperturbed model \refCite{gumbel1954statistical}.
In practice, however, models are not perturbed in the ideal form and approximations are used \refCite{papandreou2011perturb}. Hazan \textit{et al.} show that lower order approximations provide
an upper bound on the partition function \refCite{hazan2012partition}. This suggest that perturb and MAP sampling procedure can be used in the 
negative phase to maximize a lower bound on the log-likelihood of the data.
However, this is feasible only if efficient MAP estimation is possible (\eg PGMs with submodular potentials \refCite{kolmogorov2004energy}), 
and even so, repeated MAP estimation at each step of learning could be prohibitively expensive. 

Here we propose a new approach closely related to CD and perturb and MAP to sample the PGM in the negative phase of learning.
The basic idea is to perturb the model and starting at training data, find lower perturbed-energy configurations. Then use
these configurations as fantasy particles in the negative phase of learning.
Although this scheme may be used for arbitrary discrete PGMs with and without hidden variables, here we consider its application to 
the task of training Restricted Boltzmann Machine (RBM) \refCite{smolensky1986information}.

\section{Background}
\subsection{Restricted Boltzmann Machine}

RBM is a bipartite Markov Random Field, where the variables $\xv = \{ \vv, \hv\}$ 
are partitioned into visible $\vv = [v_1,\ldots,v_n]$ and hidden $\hv = [h_1,\ldots,h_m]$ units.
Because of its representation power (\refCite{le2008representational}) and relative ease of training \refCite{hinton2010practical}, RBM is increasing used in 
various applications. For example as generative model for movie ratings \refCite{salakhutdinov2007restricted}, speech \refCite{mohamed2010phone} and topic modeling \refCite{hinton2009replicated}.
Most importantly it is used in construction of deep neural architectures \refCite{hinton2006fast, bengio2009learning}.

RBM models the joint distribution over hidden and visible units by
 \begin{align*}
 \p(\hv, \vv | \theta) = \frac{1}{\Z(\theta)} e^{-E(\hv, \vv, \theta)}
 \end{align*}
where $\Z(\theta) = \sum_{\hv,\vv} e^{-E(\hv, \vv, \theta)} $ is the normalization constant (\aka partition function) and $E$ is the energy function.
Due to its bipartite form, conditioned on the visible (hidden) variables the hidden (visible) variables in an RBM are independent of each other:
\begin{align}\label{eq:independent}
\p(\hv | \vv, \theta) = \prod_{1 \leq j \leq m} \p(h_j | \vv, \theta) \quad \text{and} \quad
\p(\vv | \hv, \theta) = \prod_{1 \leq i \leq n} \p(v_i | \hv, \theta)
\end{align}

Here we consider the energy function of binary RBM, where $ h_j,v_i \in \{0,1\}$
\begin{align*}
E(\vv,\hv, \theta) \; = \;  - \bigg( \sum_{\substack{1 \leq i \leq n \\ 1 \leq j \leq m}} v_i W_{i,j} h_j 
+\sum_{1 \leq i \leq n} a_i v_i + \sum_{1 \leq j \leq m} b_j h_j \bigg )  \; = \;  
- \bigg ( \vv^{T} \W \hv + \av^T \vv + \bv^T \hv \bigg ) 
\end{align*}

The model parameter $\theta = (\W, \av, \bv)$, consists of the  matrix of $n \times m$ real valued pairwise interactions $\W$,
and local fields (\aka bias terms) $\av$ and $\bv$. The marginal over visible units is
\begin{align*} 
\p(\vv| \theta) = \frac{1}{\Z(\theta)} \sum_{\hv} \p(\vv,\hv |\theta)
\end{align*}
Given a data-set $\set{D} = \{ \vvd{1},\ldots,\vvd{N}\}$, 
maximum-likelihood learning of the model seeks the maximum of the averaged log-likelihood:
\begin{align} 
\ell(\theta) &= \frac{1}{N} \sum_{\vvd{k} \in \set{D}}  \log( \p(\vvd{k} | \theta)) \\
			&= -\frac{1}{N} \sum_{\vvd{k} \in \set{D}} \prod_{1 \leq j \leq m} \bigg ( 1 + \exp(\sum_{1 \leq i \leq n} \vd{k}_i W_{i,j})\bigg ) \; - \; \log(\Z(\theta))    			    
\end{align}

Simple calculations gives us the derivative of this objective wrt $\theta$:
\begin{align*}
\partial \ell(\theta)/\partial W_{i,j} &= \frac{1}{N} \sum_{\vvd{k} \in \set{D}} \ex{\p(h_j | \vvd{k}, \theta)} {\vd{k}_i h_j} - \ex{\p(v_i, h_j| \theta)}{v_i h_j}\\
\partial \ell(\theta)/\partial a_i &= \frac{1}{N} \sum_{\vvd{k} \in \set{D}} {\vd{k}_i} - \ex{\p(v_i| \theta)}{v_i}\\
\partial \ell(\theta)/\partial b_j &= \frac{1}{N} \sum_{\vvd{k} \in \set{D}} \ex{\p(h_j | \vvd{k}, \theta)} {h_j} - \ex{\p(h_j| \theta)}{h_j}
\end{align*}
where the first and the second terms in each line correspond to positive and negative phase respectively. 
It is easy to calculate $\p(h_j | \vvd{k}, \theta)$, required in the positive phase.
The negative phase, however, requires unconditioned samples from the current model, which may require long mixing of the Markov chain.

Note that the same form of update appears when learning any Markov Random Field, regardless of the form of graph and presence of hidden variables.
In general the gradient update has the following form
\begin{align}\label{eq:grad}
\nabla_{\theta_I} \ell(\theta) = \ex{\set{D},\theta}{\phi_I(\xv_I)} - \ex{\theta}{\phi_I(\xv_I)} 
\end{align}
where $\phi_I(\xv_I)$ is the sufficient statistics corresponding to parameter $\theta_I$. For example the sufficient statistics for 
variable interactions $W_{i,j}$ in an RBM is $\phi_{i,j}(v_i,h_j) = v_i h_j$. Note that $\theta$ in calculating the expectation of the first term appears only if
hidden variables are present.

\subsection{Contrastive Divergence Training}
In estimating the second term in the update of \refEq{eq:grad}, we can sample the model with the training data in mind. 
To this end, CD samples the model by initializing the Markov chain to data points and running it for $K$ steps. This is repeated each time we calculate the gradient. 
At the limit of $K \to \infty$, this gives unbiased samples from 
the current model, however using only few steps, CD performs very well in practice \refCite{hinton2002training}. For RBM this Markov chain is simply a block Gibbs sampler with visible and hidden units are sampled alternatively using \refEq{eq:independent}.

It is also possible to initialize the chain to the training data at the beginning of learning and during each calculation of gradient run the chain from its previous state. 
This is known as persistent CD \refCite{tieleman2008training} or stochastic maximum likelihood \refCite{younes1989parametric}.

\subsection{Sampling by Perturb and MAP}\label{sec:pmap}
Assuming that it is possible to efficiently obtain the MAP assignment in an MRF, it is possible to use 
perturbation methods to produce unbiased samples. These samples then may be used in the negative phase of learning.

Let $\Eeps(\xv) = E(\xv) - \epsilon(\xv)$ denote the perturbed energy function, 
where the perturbation for each $\xv$ is a sample from standard Gumbel distribution $\epsilon(\xv) \sim \gamma(\varepsilon) = \exp(\varepsilon - \exp(-\varepsilon))$.
Also let $\peps(\xv) \propto \exp(-\Eeps)$ denote the perturbed distribution.
Then the MAP assignment $\arg_{\xv}\max \peps(\xv)$ is an unbiased sample from $\p(\xv)$.
This means we can sample $\p(\xv)$ by repeatedly perturbing it and finding the MAP assignment.
To obtain samples from a Gumbel distribution we transform samples from uniform distribution $u \sim \set{U}(0,1)$ by  $\epsilon \leftarrow \log(-\log(u))$.

The following lemma clarifies the connection between Gibbs distribution and the MAP assignment in the perturbed model.
\begin{lemma}[\refCite{gumbel1954statistical}]\label{lemma1}
Let $\{ E(\xv)\}_{\xv \in \set{X}}$ and $E(\xv) \in \Re$. Define the perturbed values as $\Eeps(\xv) = E(\xv) - \epsilon(\xv)$, when
$\epsilon(\xv) \sim \gamma(\varepsilon) \; \forall \xv \in \set{X}$ are IID samples from standard Gumbel distribution. Then 
\begin{align}\label{eq:lemma}
Pr(\arg_{\xv \in \set{X}} \max \{ -\Eeps(\xv)\} = \xvh) = \frac{\exp(-E(\xvh))}{\sum_{\yv \in \set{X}} \exp(-E(\yv)) })
\end{align} 
\end{lemma}

Since the domain $\set{X}$, of joint assignments grows exponentially with the number of variables, we can not find the MAP assignment efficiently. 
As an approximation we may use fully decomposed noise $\epsilon(\xv) = \sum_{i} \epsilon(x_i)$ \refCite{papandreou2011perturb}. This corresponds to adding a Gumbel noise to 
each assignment of unary potentials. In the case of RBM's parametrization, this corresponds to adding the difference of two random samples from a standard Gumbel distribution (which is basically
a sample from a \emph{logistic distribution}) to biases (\eg $\aeps_i = a_i + \epsilon(v_i = 1) - \epsilon(v_i = 0)$).
Alternatively a second order approximation may perturb a combination of binary and unary potentials such that each variable is included once (\refSection{sec:second})

\section{Perturb and Descend Learning}
Feasibility of sampling using perturb and MAP depends on availability of efficient optimization procedures.
However MAP estimation is in general NP-hard \refCite{shimony1994finding} and only
a limited class of MRFs allow efficient energy minimization \refCite{kolmogorov2004energy}.
We propose an alternative to perturb and MAP that is suitable when inference is employed within the context of learning.
Since first and second order perturbations in perturb and MAP, upper bound the partition function \refCite{hazan2012partition},
 likelihood optimization using this method is desirable (\eg \refCite{wainwright2008graphical}). On the other hand since the model is trained on a data-set, 
 we may leverage the training data in sampling the model.

Similar to CD at each step of the gradient we start from training data. In order to produce fantasy particles of the negative phase we perturb the current model
and take several steps towards lower energy configurations. We may take enough steps to reach a local optima or stop midway.

Let $\thetaeps = (\Weps, \aveps, \bveps)$ denote the perturbed model. For RBM, each step of this block coordinate descend takes the following form
\begin{align}
\vv \;& \leftarrow \; \aveps + \Weps \hv > \mathbf{0}\\
\hv \;& \leftarrow \; \bveps + \Weps ^{T} \vv > \mathbf{0}
\end{align}
where starting from $\vv = \vvd{k} \in \set{D}$, $\hv$ and $\vv$ are repeatedly updated for $K$ steps or until the update above has no effect (\ie a local optima is reached).
The final configuration is then used as the fantasy particle in the negative phase of learning. 

\subsection{Amount of Perturbations}
To see the effect of the amount of perturbations we simply multiplied the  noise $\epsilon$ by a constant $\beta$ --
\ie $\beta > 1$ means we perturbed the model with larger noise values.
Going back to Lemma1, we see that any multiplication of noise can be compensated by a change of temperature 
of the energy function -- \ie for $\beta = \frac{1}{T}$, the $\arg_{\xv} \max  \Eeps(\xv) = \arg_{\xv} \max \frac{1}{T} E(\xv) - \beta \epsilon(\xv)$
remains the same. However here we are only changing the noise without changing the energy.

Here we provide some intuition about the potential effect of increasing perturbations. 
Experimental results seem to confirm this view.
For $\beta > 0$, in the negative phase of learning, we are lowering the probability of configurations 
that are at a ``larger distance'' from the training data, compared to training with $\beta = 1$.
This can make the model more robust as it puts more effort into removing false valleys that
are distant from the training data, while less effort is made to remove (false) valleys that are closer to
the training data. 



\subsection{Second Order Perturbations for RBM}\label{sec:second}
As discussed in \refSection{sec:pmap} a first order perturbation of $\theta$, only injects noise to local potentials:
\begin{align*}
\aeps_i &= a_i + \epsilon(v_i = 1) - \epsilon(v_i = 0) \quad \text{and} \quad \beps_i = b_j + \epsilon(h_i = 1) - \epsilon(h_i = 0)
\end{align*}

In a second order perturbation we may perturb a subset of non-overlapping pairwise potentials as well as unary potentials over
the remaining variables. In doing so it is desirable to select the pairwise potentials with higher influence
-- \ie larger $|W_{i,j}|$ values. With $n$ visible and $m$ hidden variables, we can use Hungarian maximum bipartite matching algorithm
to find $\min(m,n)$ most influential interactions \refCite{munkres1957algorithms}.

Once influential interactions are selected, we need to perturb the corresponding $2 \times 2$ factors with Gumbel noise as well as the bias terms for all the 
variables that are not covered. 
A simple calculation shows that perturbation of the $2 \times 2$ potentials in RBM corresponds to perturbing $W_{i,j}$ as well as $a_i$ and $b_j$ as follows
\begin{align*}
\Weps_{i,j} &= \W_{i,j} + \epsilon(1,1) - \epsilon(0,1) - \epsilon(1,0) + \epsilon(0,0)\\
\aeps_{i} &= a_i - \epsilon(0,0) + \epsilon(0,1)\\
\beps_{j} &= b_j - \epsilon(0,0) + \epsilon(1,0) \quad \quad \epsilon(0,0),\epsilon(0,1),\epsilon(1,0),\epsilon(1,1) \sim \gamma(\varepsilon)
\end{align*}
where $\epsilon(y,z)$ is basically the injected noise to the pairwise potential assignment for $v_i = y$ and $h_j = z$.

\bibliography{document}

\bibliographystyle{IEEEtran.bst}

\end{document}